\documentclass{article}
\usepackage[main, final]{neurips_2026}
\setcitestyle{square,comma,numbers,compress}
\usepackage[utf8]{inputenc}
\usepackage[T1]{fontenc}
\usepackage{amsmath,amssymb,amsfonts}
\usepackage{graphicx}
\usepackage{booktabs}
\usepackage{hyperref}
\usepackage{url}
\usepackage{xcolor}
\usepackage{colortbl}
\usepackage{subcaption}
\usepackage{multirow}
\usepackage{enumitem}
\usepackage{nicefrac}
\usepackage{float}

\title{TopoCurve: Geometry-Aware Topology Reasoning \\ via B\'{e}zier Curves in Autonomous Driving}

\author{%
  Mihai Bogdan Deaconu \\
  Bosch Engineering Center Cluj, Cluj-Napoca, Romania \\
  Faculty of Mathematics and Computer Science\\
  Babeș-Bolyai University, Cluj-Napoca, Romania \\
  \texttt{mihai.deaconu@stud.ubbcluj.ro} \\
  \And
  Laura Dioșan \\
  Faculty of Mathematics and Computer Science\\
  Babeș-Bolyai University, Cluj-Napoca, Romania \\
  \texttt{laura.diosan@ubbcluj.ro} \\
}

\begin{document}
\maketitle

\begin{abstract}
Topology reasoning jointly detects 3D lanes and traffic elements from multi-view images and infers their structural connectivity.
Current methods model lanes as discrete polylines, lacking smoothness, analytical tangent directions, and global spatial support for attention, while providing sparse topology supervision.
We propose TopoCurve, a geometry-driven architecture for 3D topology reasoning grounded in a structured parametric lane representation.
Lanes are modeled as endpoint-fixed cubic Bézier curves, enabling continuous geometry with exact endpoints and analytically defined directionality.
We exploit this shared curve geometry across the entire pipeline.
Endpoint distance and tangent alignment are encoded with multi-scale Fourier features and injected into the topology head.
Sampled curve points serve as geometry-aligned references for deformable cross-attention spanning the full lane.
Parallel curve-anchored attention branches provide diverse predictions for one-to-many topology supervision.
These components form a tightly coupled cascade where representation enables geometric reasoning, guides feature aggregation, and supports denser supervision.
TopoCurve achieves 50.6 OLS on the OpenLane-V2 benchmark without any post-processing, establishing a new state-of-the-art among end-to-end camera-only methods, and outperforms all existing approaches on endpoint detection (56.8 vs.\ 52.6 on DET$_p$).
\end{abstract}
 
\section{Introduction}
\label{sec:intro}

Accurate perception of lane markings and traffic elements is essential for understanding complex road environments in autonomous driving.
Beyond detecting individual elements, the system must infer how lanes connect and which traffic elements govern which lanes.
This joint task, known as topology reasoning~\cite{wang2023openlane,li2023toponet,fu2024topologic,fu2025topopoint}, has become central to end-to-end driving systems~\cite{hu2023planning,casas2021mp3} and HD map construction~\cite{li2022hdmapnet,liu2023vectormapnet,liao2022maptr}.
The query-based architecture is the mainstream approach: lanes, traffic elements, and endpoints are encoded through learnable queries paired for topology prediction.
However, the dominant lane representation of 11 discrete 3D points introduces structural limitations.

\textbf{Limitations of the polyline representation.}
Predicting points independently lacks smoothness constraints, allowing jagged, self-intersecting or physically implausible shapes.
Methods that observe this problem resort to inference-time heuristics, rather than addressing the representation itself: TopoPoint~\cite{fu2025topopoint} proposes Point-Lane Geometry Matching (PLGM), snapping endpoints to detected point queries, gaining 0.5 OLS via post-processing alone.
Beyond smoothness, independent coordinates carry no geometric inductive bias and cannot express curvature or directional continuity, being poorly suited for
downstream trajectory planning which benefits from continuous parametric curves.
Topology is fundamentally determined at lane endpoints and by directional compatibility: two lanes with nearby endpoints may diverge in opposite directions at forking intersections.
Polylines lack stable analytical tangents at these critical locations, and computing them via finite differences on discrete points is numerically noisy and not differentiable through the matching process.

\textbf{Limitations of existing topology reasoning.}
Methods that rely solely on dot-product feature similarity~\cite{wu2024topomlp} lack explicit geometric reasoning about directional alignment.
TopoLogic~\cite{fu2024topologic} combines geometric distance with query similarity and injects both into the GCN adjacency, coupling topology training to feature propagation and potentially degrading detection quality.
Raw distance scalars suffer from spectral bias~\cite{tancik2020fourier}: the boundary between connected and unconnected lanes is a high-frequency function that standard MLPs approximate poorly.
Furthermore, standard deformable attention~\cite{zhu2021deformable} samples around a single centroid, covering only a fraction of 30--50\,m lanes and leaving critical endpoints unobserved.
Finally, one-to-one Hungarian matching with typically ${\sim}$10 ground truth lanes and 300 queries produces sparse valid topology supervision. 
Together, these limitations hinder both the geometric fidelity of lane representations and the accuracy of topology reasoning.

\textbf{Our approach.}
We propose TopoCurve, a geometry-driven architecture for 3D lane topology reasoning, centered on a structured lane representation. We parameterise lanes as endpoint-fixed cubic B\'{e}zier curves, providing compact, smooth geometry with exact endpoints. 
This representation enables the extraction of geometric cues (distance and directional alignment), supports geometry-aligned feature aggregation through curve-based attention, and facilitates redundancy supervision through multiple curve-anchored predictions.
While each component contributes independently, they function most effectively as a cascade where each builds on the preceding one (Tables~\ref{tab:incremental},~\ref{tab:leaveout}), forming a geometry-driven pipeline where representation, feature aggregation, and supervision are tightly coupled. Our contributions are summarized as follows:

\begin{enumerate}[leftmargin=*,nosep]
    \item We introduce 
    endpoint-fixed cubic B\'{e}zier curves for 3D lane topology reasoning, reducing the regression target from 33 to 12 dimensions with guaranteed smoothness, deterministic tangent directions, and exact endpoint localization, addressing the geometric inconsistencies of polyline representations
    (Section~\ref{sec:bezier}). This representation enables the other geometry-aware components within a single coherent pipeline.
    \item We propose FourierGeoBias, a geometry injection mechanism that extracts analytical tangent directions and endpoint distances from B\'{e}zier control points, encodes them with multi-scale Fourier features to mitigate spectral bias, and injects the result into the topology head, enabling topology prediction with explicit geometric reasoning beyond feature similarity (Section~\ref{sec:fourier}).
    \item We design B\'{e}zier-Distributed Deformable Attention (BDDA), replacing centroid-based sampling with curve-anchored references that provide a full-lane receptive field, improving feature aggregation for long and spatially extended lanes (Section~\ref{sec:bdda}).
    \item We propose Curve-Anchored Redundancy Supervision (CARS), parallelizing BDDA into multiple instances for one-to-many topology supervision with FourierGeoBias, mitigating the sparsity of one-to-one Hungarian matching, with the one-to-many supervision adding no inference cost (Section~\ref{sec:cars}).
\end{enumerate}

TopoCurve achieves state-of-the-art results on both OpenLane-V2 subsets (50.6 OLS), demonstrating cross-dataset generalization. Furthermore, we outperform all existing methods on TopoPoint’s proposed DET$_p$ endpoint metric (56.8 vs.\ 52.6) without post-processing heuristics (Section~\ref{sec:experiments}).
\section{Related Work}
\label{sec:related}

\textbf{Lane representations.}
The dominant paradigm models each lane as a polyline of $K$ discrete 3D points~\cite{li2023toponet,wu2024topomlp,fu2024topologic,fu2025topopoint,lv2025topoformer}.
This representation is neither continuous nor structurally constrained, lacks analytical access to geometric properties such as tangents, and requires high-dimensional regression.
Sparse BEV-based detectors such as DETR3D~\cite{wang2021detr3d} and PETR~\cite{liu2022petr} form the backbone of these approaches, while PersFormer~\cite{chen2022persformer}, Anchor3DLane~\cite{huang2023anchor3dlane}, CurveFormer~\cite{bai2023curveformer}, LATR~\cite{luo2023latr}, and LaneSegNet~\cite{li2024lanesegnet} improve geometric awareness at the feature, anchor, query, or segment level without altering the underlying polyline representation.
TopoNet~\cite{li2023toponet}, TopoMLP~\cite{wu2024topomlp}, TopoLogic~\cite{fu2024topologic}, TopoPoint~\cite{fu2025topopoint}, and TopoFormer~\cite{lv2025topoformer} extend this paradigm with different topology heads and architectural choices.
Parametric alternatives, including B\'{e}zier curves, B-splines~\cite{pittner2024lanecpp}, and polynomial models~\cite{feng2022bezier,chougule2023bsnet,tabelini2021polylane}, provide continuous and differentiable representations.
However, they are typically unconstrained: endpoints are not enforced, supervision is often applied in parameter space, and control points lack direct geometric interpretability.
B\'{e}zierLaneNet~\cite{feng2022bezier}, BSNet~\cite{chougule2023bsnet}, and PolyLaneNet~\cite{tabelini2021polylane} follow this paradigm, while TopoMLP~\cite{wu2024topomlp} adopts a similar parameterisation without integrating it into downstream topology reasoning.
Despite these advances, structurally constrained parametric representations have not been explored for 3D lane topology reasoning.

\textbf{Geometry in topology reasoning.}
Early approaches rely on heuristic or distance-based signals~\cite{can2022stsu,liao2023lanegap}, while recent methods integrate topology prediction into end-to-end architectures.
However, geometric cues are often implicit (embedded in features) or limited to low-dimensional scalars such as pairwise distances~\cite{wu2024topomlp,fu2024topologic,fu2025topopoint,lv2025topoformer}.
These signals are typically encoded as raw or single-scale functions, making them prone to spectral bias~\cite{tancik2020fourier}.
Moreover, geometric priors are often injected into intermediate stages such as graph adjacency~\cite{li2023toponet,fu2024topologic}, coupling topology reasoning with feature propagation and potentially degrading detection.
A key limitation is the absence of explicit directional reasoning: no approach provides analytical tangents at lane endpoints.
Our FourierGeoBias addresses these issues by encoding distance and tangent alignment with multi-scale Fourier features and injecting them directly into the prediction head, decoupling reasoning from feature learning.

\textbf{Deformable attention for elongated structures.}
Standard deformable attention~\cite{zhu2021deformable} samples around a single centroid, which is inadequate for 30--50\,m lanes because sampling is learned implicitly without structural geometric constraints. BEVFormer~\cite{li2022bevformer} improves spatial consistency, but retains centroid-based sampling. RelTopo~\cite{luo2025reltopo} partially addresses this limitation by sampling features along predicted B\'{e}zier curves in BEV, but uses unconstrained B\'{e}zier references and is not designed for parallelisation or connection to downstream topology supervision. Our BDDA introduces geometry-explicit attention in BEV space using endpoint-fixed references. These are progressively refined across layers via the Bernstein basis used for regression, serving as CARS's building block.

\textbf{One-to-many assignment for topology.}
Group DETR~\cite{chen2023groupdetr} and Hybrid Matching~\cite{jia2023hdetr} introduced O2M detection branches.
RATopo~\cite{li2025ratopo} adapts O2M for topology by reordering the decoder (cross-attention before self-attention) and instantiating parallel cross-attention blocks.
However, RATopo uses centroid-based standard deformable attention (not curve-distributed), dot-product topology heads (no geometric priors), polyline matching, and does not account for a coupled point-lane decoder with GCN message passing.
Our CARS addresses all of these: each parallel block is a curve-anchored BDDA instance, O2M includes FourierGeoBias, matching operates in B\'{e}zier-projected space, and the mechanism integrates with a joint point-lane decoder and scene graph network.

\section{Method}
\label{sec:method}

\subsection{Problem Definition}
\label{sec:problem}
Given images from $N_\text{cam}$ cameras, topology reasoning involves 3D lane centerline detection~\cite{chen2022persformer,huang2023anchor3dlane} in BEV, 2D traffic element detection~\cite{carion2020detr} in front-view, and inferring their connectivity~\cite{li2023toponet,wang2023openlane}. 
Formally, we detect $N_l$ lanes as sequences of $K$ points $\mathcal{L} = \{L_i \in \mathbb{R}^{K \times 3}\}_{i=1}^{N_l}$ and $N_t$ traffic elements as 2D boxes $\mathcal{T} = \{T_j \in \mathbb{R}^{4}\}_{j=1}^{N_t}$. Their interactions are captured by binary adjacency matrices: $\mathbf{G}^{ll} \in \{0,1\}^{N_l \times N_l}$ for lane-lane traversability and $\mathbf{G}^{lt} \in \{0,1\}^{N_l \times N_t}$ for lane-element assignment. Following~\cite{fu2025topopoint}, we additionally predict $N_p$ unique lane endpoints $\mathcal{P} = \{P_k \in \mathbb{R}^{3}\}_{k=1}^{N_p}$, linking them to their corresponding lanes via a point-lane topology $\mathbf{G}^{pl} \in \{0,1\}^{N_p \times N_l}$.

\subsection{Architecture Overview}
\label{sec:overview}

\begin{figure}[t]
\centering
\includegraphics[width=\textwidth]{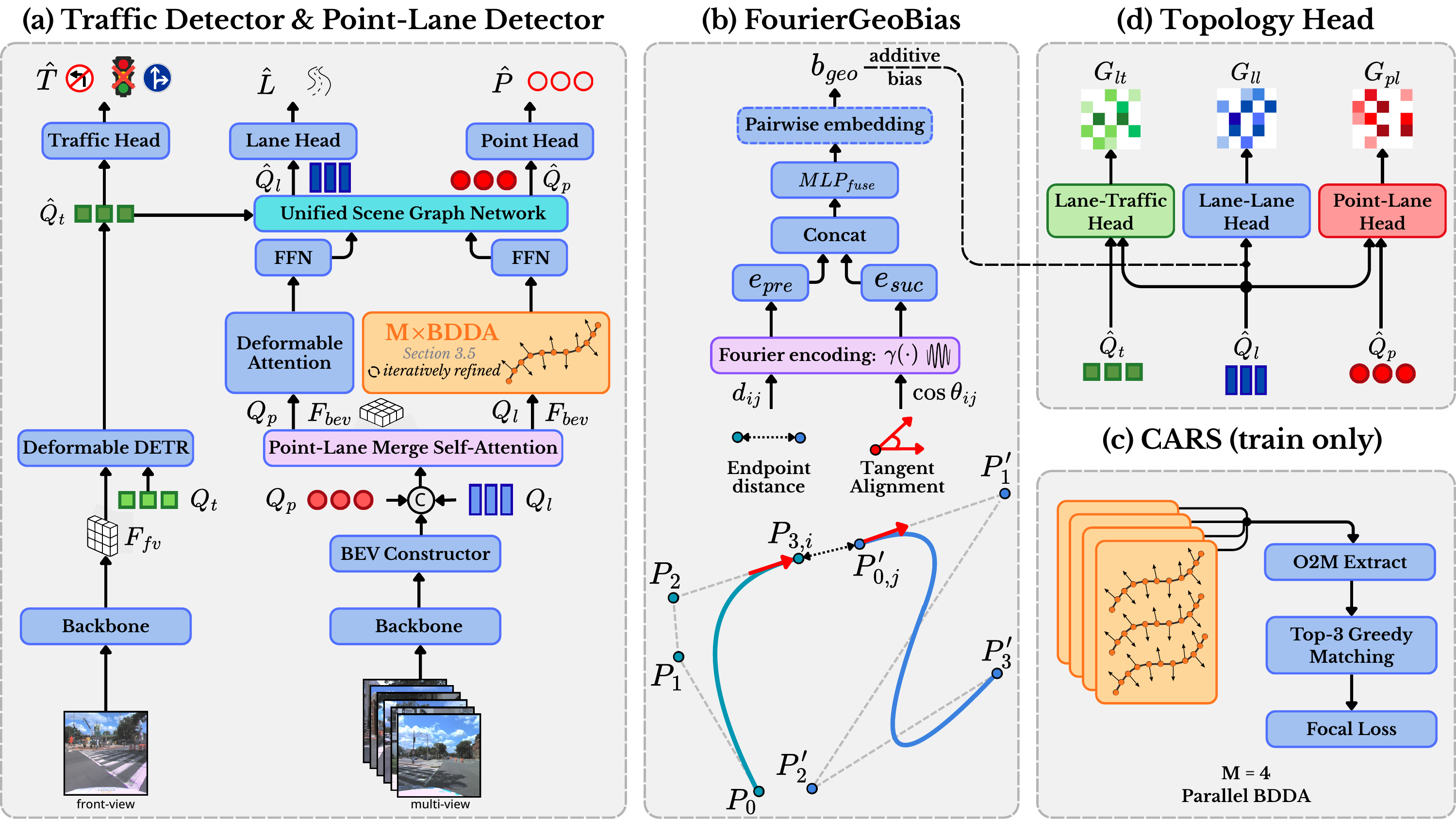}
\caption{\textbf{TopoCurve architecture.} (a)~The point-lane decoder processes queries through 6 layers, each containing 4 parallel BDDA blocks for lanes and standard deformable attention for points, alongside a standard Deformable DETR head for traffic elements. (b)~FourierGeoBias injects Fourier-encoded distance and tangent into the lane-lane topology head. (c)~During training, CARS extracts pre-fusion queries for O2M topology supervision. (d)~Topology heads produce predictions.
}
\label{fig:architecture}
\end{figure}

As illustrated in Figure~\ref{fig:architecture}, TopoCurve builds on a query-based topology reasoning pipeline in which lanes, endpoints, and traffic elements are represented as learnable queries, enabling joint detection and topology prediction.
Multi-view images are processed by a shared ResNet-50~\cite{he2016resnet} backbone with FPN~\cite{lin2017fpn} to produce multi-scale features, which are encoded using BEVFormer~\cite{li2022bevformer}.
Traffic element queries $\mathbf{Q}_t$ interact with front-view features through a Deformable DETR~\cite{zhu2021deformable} head for 2D bounding box prediction.
The core of TopoCurve is the \textbf{point-lane decoder}, which jointly processes lane queries $\mathbf{Q}_l$ and endpoint queries $\mathbf{Q}_p$ across $L{=}6$ transformer layers.
At each layer, lane queries attend to BEV features through parallel BDDA blocks (Section~\ref{sec:bdda}), while point queries use standard deformable cross-attention with independently initialized reference points.
The two query sets are then concatenated along the instance dimension and processed by joint multi-head self-attention, enabling global information exchange between points and lanes before being split back. The resulting queries are passed through the Unified Scene Graph Network (USGN)~\cite{fu2025topopoint}, a two-round graph convolution module~\cite{kipf2017gcn} that performs bidirectional message passing between point, lane, and traffic element representations using distance-derived adjacency matrices.
Lane queries predict B\'{e}zier control points (Section~\ref{sec:bezier}). Topology heads (Section~\ref{sec:topo_head}) compute connectivity with FourierGeoBias (Section~\ref{sec:fourier}), and during training, CARS (Section~\ref{sec:cars}) extracts pre-fusion queries for one-to-many topology supervision.

\subsection{Endpoint-Fixed B\'{e}zier Reparameterization}
\label{sec:bezier}
In standard topology reasoning pipelines, each lane query regresses $K{=}11$ independent 3D points (33 dimensions) with no structural coupling.
The proposed reparameterisation transforms the lane from a discrete set of points into a continuous geometric object with explicit structure, making geometric reasoning for topology directly accessible. 
We reparameterize each lane as a cubic B\'{e}zier curve with 4 control points $\mathbf{C} = [\mathbf{P}_0, \mathbf{P}_1, \mathbf{P}_2, \mathbf{P}_3] \in \mathbb{R}^{4 \times 3}$, reducing the target to 12 dimensions (63.6\% reduction) with guaranteed $C^\infty$ smoothness:
\begin{equation}
\mathbf{B}(t) = \textstyle\sum_{j=0}^{3} \binom{3}{j} (1{-}t)^{3-j} t^j \mathbf{P}_j, \quad t \in [0,1]
\label{eq:bezier}
\end{equation}
While recent works~\cite{pittner2024lanecpp} use B-splines, they are not a suitable parameterization for topology, because their first and last control points are not guaranteed to align exactly with the centerline's endpoints. B\'{e}zier curves mathematically guarantee this alignment, an essential property we require.

\textbf{Matching and loss in projected space.}
B\'{e}zier parametric sampling at fixed $t_k$ does not produce arc-length equidistant points: samples cluster in high-curvature regions.
Since OpenLane-V2 provides ground-truth lanes as $K{=}11$ approximately arc-length equidistant 3D points, directly comparing parametric samples against this ground truth would incur nonzero cost even for perfect predictions.
This mismatch necessitates a consistent sampling strategy for both predictions and targets.

\textbf{Endpoint-fixed ground-truth fitting.}
To obtain a consistent B\'{e}zier representation, we fit a cubic curve to the ground truth points while enforcing exact endpoint alignment: $\mathbf{P}_0$ and $\mathbf{P}_3$ are pinned to the first and last points, and only the interior control points $\mathbf{P}_1, \mathbf{P}_2$ are solved via least-squares:
\begin{equation}
[\mathbf{P}_1^*, \mathbf{P}_2^*] = (\mathbf{A}_\text{int}^\top \mathbf{A}_\text{int})^{-1} \mathbf{A}_\text{int}^\top ( \mathbf{Y}_\text{int} - \mathbf{A}_{[:,0]} \mathbf{P}_0 - \mathbf{A}_{[:,-1]} \mathbf{P}_3)
\label{eq:fitting}
\end{equation}
where $\mathbf{A} \in \mathbb{R}^{K \times 4}$ is the Bernstein basis matrix with entries $A_{kj} = \binom{3}{j}(1{-}t_k)^{3-j}t_k^j$ at $t_k = k/(K{-}1)$, $\mathbf{Y}_\text{int}$ contains interior ground truth points, and subscript ``int'' indexes interior rows and columns.
By fitting ground truth to control points and re-sampling both predictions and targets through the same Bernstein basis, both share the same parametric pattern, and the matching cost directly reflects true geometric similarity.
The denser high-curvature sampling further induces implicit curvature-adaptive weighting in the L1 loss.
Endpoint-fixing ensures exact ground truth endpoint preservation, which is essential for topology where connectivity is defined at lane boundaries.

\textbf{Bernstein basis projection.}
We precompute $\mathbf{A} \in \mathbb{R}^{K \times 4}$ as a shared buffer.
Predicted control points $\hat{\mathbf{C}}$ project to $K$ sample points via $\hat{\mathbf{L}} = \mathbf{A}\hat{\mathbf{C}} \in \mathbb{R}^{K \times 3}$, serving three roles: lane point recovery for matching and loss, curve reference computation for BDDA, and coordinate generation for USGN distance computation.
Matching and L1 regression loss operate in this 33-dimensional projected space, ensuring assignment cost is not distorted by the nonlinear control-point-to-curve mapping.

While a single cubic B\'{e}zier is theoretically limited for S-curves, OpenLane-V2 represents complex geometries as sequences of connected simpler segments, making cubic parameterization a natural match.
Empirical evaluations confirm $n_\text{ctrl}{=}4$ as optimal~\cite{wu2024topomlp,feng2022bezier}, with discrete Fr\'{e}chet distance validation yielding median 0.011\,m and mean 0.086\,m reconstruction error across the training set.

\subsection{Fourier-Enhanced Directional Geometry Injection}
\label{sec:fourier}
The B\'{e}zier control points provide analytical access to endpoint distances and tangent directions, geometric signals unavailable from polylines.
We encode these features with multi-scale Fourier encoding and inject them as an additive bias into the topology prediction head, decoupling geometric reasoning from feature learning without degrading detection.

\textbf{Geometric features from B\'{e}zier control points.}
Given predicted control points $\hat{\mathbf{C}}_i, \hat{\mathbf{C}}_j$, we extract the endpoint distance $d_{ij}$ and tangent alignment $\cos\theta_{ij}$:
\begin{equation}
d_{ij} = \|\mathbf{P}_{3,i} {-} \mathbf{P}_{0,j}\|_2, \\ \cos\theta_{ij} = \hat{\mathbf{t}}_{3,i} {\cdot} \hat{\mathbf{t}}_{0,j}, \; \hat{\mathbf{t}}_{3,i} = \frac{\mathbf{P}_{3,i} {-} \mathbf{P}_{2,i}}{\|\mathbf{P}_{3,i} {-} \mathbf{P}_{2,i}\|}, \; \hat{\mathbf{t}}_{0,j} = \frac{\mathbf{P}_{1,j} {-} \mathbf{P}_{0,j}}{\|\mathbf{P}_{1,j} {-} \mathbf{P}_{0,j}\|}
\label{eq:geo_features}
\end{equation}
where $d_{ij}$ measures the Euclidean distance between lane $i$'s endpoint and lane $j$'s startpoint, and $\cos\theta_{ij}$ captures directional compatibility.
The tangent directions are exact analytical B\'{e}zier derivatives at $t{\in}\{0,1\}$, requiring no finite-difference approximation.
Intuitively, because connected centerlines must flow in the same direction at their junction, tangent alignment provides a powerful discriminative signal: connected lane pairs in our training data exhibit $>$97\% average tangent cosine similarity, far exceeding that of unconnected pairs at similar distances.

\textbf{Directional pairwise embedding.}
Lane connectivity is directional, depending on where one lane ends and another begins. To model this asymmetry, we compute separate predecessor and successor embeddings.
All MLPs in FourierGeoBias are two-layer networks that compress representations to 64 dimensions.
For the lane queries and corresponding endpoint coordinates, we apply sinusoidal positional encoding (SinPE), where $\mathbf{q}_i$ is the lane query and $\text{proj}$ is a learnable linear projection:
\begin{align}
\mathbf{e}_{\text{pre},i} &= \text{MLP}_\text{pre}(\mathbf{q}_i) + \text{proj}_\text{pre}(\text{SinPE}(\mathbf{P}_{3,i})) \label{eq:pre} \\
\mathbf{e}_{\text{suc},j} &= \text{MLP}_\text{suc}(\mathbf{q}_j) + \text{proj}_\text{suc}(\text{SinPE}(\mathbf{P}_{0,j})) \label{eq:suc}
\end{align}

For the explicit distance and tangent alignment, we apply multi-scale Fourier encoding $\gamma(\cdot)$:
\begin{equation}
\mathbf{e}_{\text{geo}}[i,j] = \text{MLP}_\text{dist}\big([\gamma(d_{ij}/50) \;\|\; \gamma(\cos\theta_{ij})]\big) \label{eq:geo}
\end{equation}
Because raw scalars are difficult for MLPs to model, we encode them using Fourier features~\cite{tancik2020fourier} to capture high-frequency variations. Each scalar $x$ is encoded as:
\begin{equation}
\gamma(x) = \big[x,\; \sin(\pi 2^0 x),\; \cos(\pi 2^0 x),\; \ldots,\; \sin(\pi 2^{L-1} x),\; \cos(\pi 2^{L-1} x)\big] \in \mathbb{R}^{2L+1}
\label{eq:fourier}
\end{equation}
with $L{=}8$ frequency bands providing sensitivity from sub-meter to over 50 meters, overcoming the spectral bias of raw-scalar MLPs.
Finally, we fuse the semantic and geometric information by concatenating these structural representations and projecting them to a scalar bias:
\begin{equation}
b_\text{geo}[i,j] = \text{MLP}_\text{fuse}\big([\mathbf{e}_{\text{pre},i} \;\|\; \mathbf{e}_{\text{suc},j} \;\|\; \mathbf{e}_{\text{geo}}[i,j]]\big)
\label{eq:bias}
\end{equation}
where MLP$_\text{fuse}$ has its last layer zero-initialized, ensuring $b_\text{geo}{=}0$ at training start.
The bias is added to the base dot-product logit: $\hat{G}^{ll}_{ij} = \mathbf{e}^s_i \cdot \mathbf{e}^s_j + b_\text{geo}[i,j]$, where $\mathbf{e}^s_i$ and $\mathbf{e}^s_j$ denote the semantically enriched embeddings of lanes $i$ and $j$ produced by the topology MLP head after graph-based feature aggregation.
FourierGeoBias applies only to the lane-lane head, while lane-traffic uses the standard dot-product (tangent alignment is meaningless for lane-to-traffic pairs).
The USGN's graph convolution (GCN) adjacency matrices remain distance-based, preserving detection quality by preventing geometry signals from interfering with feature propagation (Table~\ref{tab:fourier}).

\subsection{B\'{e}zier-Distributed Deformable Attention}
\label{sec:bdda}

Standard deformable attention samples around a single centroid per query, leaving the endpoints of 30--50\,m lanes outside the effective receptive field.
Our B\'{e}zier representation naturally provides $K{=}11$ geometry-aligned reference locations along each lane via the Bernstein basis, enabling attention to capture the full spatial extent.
The proposed BDDA replaces centroid-based sampling with these curve-distributed references, yielding a geometry-aligned receptive field spanning the full lane: 
\begin{equation}
\text{BDDA}(\mathbf{q}_l, \{\mathbf{p}_k\}, \mathbf{F}) = \textstyle\sum_{m=1}^{M_h} \mathbf{W}_m \big[ \sum_{k=1}^{K} \sum_{n=1}^{N} A_{mkn} \cdot \mathbf{W}'_m \mathbf{F}(\mathbf{p}_k {+} \Delta\mathbf{p}_{mkn}) \big]
\label{eq:bdda}
\end{equation}
where $M_h$ denotes the number of attention heads, $N$ offsets are learned per curve point, and $A_{mkn}$ and $\Delta\mathbf{p}_{mkn}$ are attention weights and sampling offsets generated from projections of $\mathbf{q}_l$.
The softmax normalization of $A_{mkn}$ is computed jointly across all $K {\times} N$ sampling points, letting the model learn which curve regions are most informative.
Because our B\'{e}zier is endpoint-fixed, the first and last anchors coincide exactly with the lane start and end.

\textbf{Iterative curve-reference refinement.}
At each decoder layer, predicted control points are projected through the Bernstein basis to produce $K$ updated BEV references for the next layer. At layer~0, all references collapse to the initial centroid, recovering standard deformable attention.

\subsection{Curve-Anchored Redundancy Supervision}
\label{sec:cars}
Under standard one-to-one Hungarian matching, each ground-truth lane supervises only a single predicted query.
With ${\sim}$10 GT lanes and 300 queries, fewer than 0.2\% of lane pairs receive valid topology supervision, leading to severe sparsity.
We exploit BDDA's multiple geometry-aware views to densify supervision, and propose CARS, a geometry-aware O2M strategy that supervises multiple curve-anchored predictions per lane, increasing both the quantity and diversity of topology signals.

\textbf{Parallel curve-anchored cross-attention.}
We instantiate $M{=}4$ parallel BDDA blocks at the decoder level.
Each block $m$ independently attends to BEV features using the $K$ curve references but with independent parameters, to capture diverse geometry-aware representations:
\begin{equation}
\bar{\mathbf{Q}}^{(i)}_m = \text{LN}\big(\text{BDDA}^{(i)}_m(\mathbf{Q}^{(i-1)}, \mathbf{F}_\text{bev}, \mathbf{p}^{(i-1)}_{1:K}) + \mathbf{Q}^{(i-1)}\big), \quad m = 1, \ldots, 4
\label{eq:multi_ca}
\end{equation}
The outputs are fused via $\bar{\mathbf{Q}}^{(i)} = \text{Linear}([\bar{\mathbf{Q}}^{(i)}_1 \;\|\; \cdots \;\|\; \bar{\mathbf{Q}}^{(i)}_4])$.
Unlike prior parallel attention~\cite{li2025ratopo,chen2023groupdetr} that samples around a single centroid, each block samples at $K$ physically meaningful curve positions, creating both parametric (independent weights) and geometric (curve-anchored) diversity.

The cross-attention is placed before self-attention: self-attention acts as a learned suppression mechanism analogous to non-maximum suppression, removing redundant predictions.
By placing the parallel blocks before suppression, we preserve diverse lane variants for O2M supervision.

\textbf{Pre-fusion O2M topology supervision.}
During training, we extract intermediate queries from each block before fusion and self-attention, maximizing diversity.
For each block $m$ and layer $i$: the shared regression head predicts B\'{e}zier control points, projects to 33 dimensions via the Bernstein basis, and top-3 greedy matching selects the best predictions per ground truth lane from the classification-regression cost.
Greedy matching is cheaper than repeated Hungarian matching and avoids the global optimality constraint that is counterproductive for multiple positives.
For lane-lane O2M, topology prediction includes FourierGeoBias, so auxiliary supervision is geometry-aware.
Focal loss is computed on all valid pairs. The choices of 4 parallel blocks and 3 O2M positives per ground-truth follow the empirical findings of~\cite{li2025ratopo}, which report diminishing returns beyond these values.

\subsection{Topology Head}
\label{sec:topo_head}
We predict topology as connectivity matrices $\hat{\mathbf{G}}$ from enhanced features $\hat{\mathbf{Q}}_p$, $\hat{\mathbf{Q}}_l$, $\hat{\mathbf{Q}}_t$ via separate MLPs and pairwise similarity, where $\sigma$ is the sigmoid function and $\mathbf{B}_\text{geo}$ is the FourierGeoBias matrix proposed in Eq.~\eqref{eq:bias}, applied only to $\hat{\mathbf{G}}^{ll}$:
\begin{gather}
\hat{\mathbf{G}}^{pl} = \sigma(\text{MLP}(\hat{\mathbf{Q}}_p) \cdot \text{MLP}(\hat{\mathbf{Q}}_l)^\top), \quad
\hat{\mathbf{G}}^{lt} = \sigma(\text{MLP}(\hat{\mathbf{Q}}_l) \cdot \text{MLP}(\hat{\mathbf{Q}}_t)^\top), \label{eq:topo_pl_lt} \\
\hat{\mathbf{G}}^{ll} = \sigma(\text{MLP}(\hat{\mathbf{Q}}_l) \cdot \text{MLP}(\hat{\mathbf{Q}}_l)^\top + \mathbf{B}_\text{geo}) \label{eq:topo_ll}
\end{gather}

\subsection{Training and Inference}
\label{sec:training}
The total loss combines detection and topology components:
\begin{equation}
\mathcal{L} = \lambda_t \mathcal{L}_t + \lambda_p \mathcal{L}_p + \lambda_l \mathcal{L}_l + \lambda_{pl} \mathcal{L}_{pl} + \lambda_{ll} \mathcal{L}_{ll} + \lambda_{lt} \mathcal{L}_{lt} + \lambda_\text{o2m} \mathcal{L}_\text{o2m}
\label{eq:total_loss}
\end{equation}
where $\mathcal{L}_t$, $\mathcal{L}_p$, $\mathcal{L}_l$ are detection losses and $\mathcal{L}_{pl}$, $\mathcal{L}_{ll}$, $\mathcal{L}_{lt}$ are topology losses.
Lane and point detection losses consist of focal loss~\cite{lin2017focal} for classification and L1 loss for regression (in 33-dimensional projected space for lanes).
Traffic element loss additionally includes GIoU loss~\cite{rezatofighi2019giou}.
All topology losses use focal loss.
The O2M auxiliary loss averages across all $M {\times} L$ block-layer combinations:
\begin{equation}
\mathcal{L}_\text{o2m} = \frac{1}{M \cdot L} \textstyle\sum_{i=1}^{L} \sum_{m=1}^{M} \lambda_{ll} \mathcal{L}^{ll}_{\text{o2m},i,m},
\label{eq:o2m_loss}
\end{equation}
keeping magnitude proportional to O2O losses. 
At inference, the O2M branch is disabled.
BDDA introduces negligible overhead, as reference computation is a single matrix multiply and deformable attention is fully GPU-parallelizable (Table~\ref{tab:efficiency}).
Unlike TopoPoint, no post-processing is required.

\section{Experiments}
\label{sec:experiments}

\subsection{Dataset and Metrics}
\label{sec:dataset}
\textbf{Dataset.}
We conduct experiments on the OpenLane-V2 benchmark~\cite{wang2023openlane}, a large-scale topology reasoning dataset built from Argoverse2~\cite{wilson2023argoverse2} (subset\_A, 7 cameras) and nuScenes~\cite{caesar2020nuscenes} (subset\_B, 6 cameras). Each subset contains 1,000 scenes captured at 2\,Hz, providing multi-view images and comprehensive annotations for lane centerlines, traffic elements, and both topology types.

\textbf{Metrics.}
Following the OpenLane-V2 evaluation protocol (v2.1.0), we report DET$_l$ (lane detection AP averaged over Fr\'{e}chet thresholds 1.0, 2.0, 3.0\,m), DET$_t$ (traffic element AP via IoU across categories), TOP$_{ll}$ and TOP$_{lt}$ (topology matrix similarity), and the overall score:
\begin{equation}
\text{OLS} = \tfrac{1}{4}\big[\text{DET}_l + \text{DET}_t + \sqrt{\text{TOP}_{ll}} + \sqrt{\text{TOP}_{lt}}\big].
\label{eq:ols}
\end{equation}
We additionally report DET$_p$~\cite{fu2025topopoint}, defined as $\text{DET}_p = \frac{1}{|T|}\sum_{t \in T} \text{AP}_t$ over thresholds $T = \{1.0, 2.0, 3.0\}$ using point-wise Fr\'{e}chet distance, to quantify endpoint detection quality.

\subsection{Implementation Details}
\label{sec:impl}

The multi-view images have a native resolution of $2048{\times}1550$ and are downsampled by a factor of $0.5$. The front view is retained at its original resolution and padded to match. We use a pretrained ResNet-50 backbone with a Feature Pyramid Network neck to produce multi-scale features with a hidden dimension of 256. The BEVFormer encoder constructs a bird's-eye-view grid of $200{\times}100$ cells. We initialise 300 lane queries, 200 point queries, and 100 traffic element queries. The point-lane decoder consists of 6 transformer layers, each containing 4 parallel BDDA blocks with 4 sampling offsets per curve point and 8 attention heads. FourierGeoBias employs 8 frequency bands and projects the fused embeddings to 64 dimensions. For CARS, the O2M matching selects 3 positives per ground-truth lane, and the USGN edge weight is set to 0.6. To ensure fair comparisons and strictly isolate the performance gains of our architecture, we adopt the general hyperparameter and loss weighting configuration of the baseline~\cite{fu2025topopoint}. Loss weights are set as $\lambda_l{=}1.5$, $\lambda_t{=}\lambda_p{=}1$, $\lambda_{pl}{=}\lambda_{ll}{=}\lambda_{lt}{=}5$, and $\lambda_\text{o2m}{=}2$. TopoCurve is trained end-to-end for 24 epochs using the AdamW optimizer with a cosine annealing learning rate
schedule, starting at $3{\times}10^{-4}$, with a weight decay of 0.01, on two NVIDIA H200 GPUs with a batch size of 12. Inference classification thresholds are set to 0.3. 

\subsection{Comparison with Current State of the Art}
\label{sec:main_results}
As shown in Table~\ref{tab:main}, TopoCurve establishes a new state-of-the-art on the OpenLane-V2 benchmark. On subset\_A, it outperforms TopoFormer (50.6 vs.\ 46.3 OLS), despite their use of a pretrained lane detector.
Compared to TopoPoint, our geometric formulation yields substantial gains in lane detection (33.9 vs.\ 31.4 DET$_l$) and topology prediction (30.7 vs.\ 28.7 TOP$_{ll}$, 33.0 vs.\ 30.0 TOP$_{lt}$).
Critically, our representation resolves endpoint localization end-to-end, surpassing TopoPoint's results without post-processing (56.8 vs.\ 52.6 on DET$_p$). Furthermore, TopoCurve achieves state-of-the-art performance on subset\_B (51.2 OLS and 48.3 DET$_p$), confirming robust cross-sensor generalization.

\begin{table}[t]
\centering
\caption{\textbf{Results on OpenLane-V2.} All methods use ResNet-50 backbone and are trained for 24 epochs. TopoFormer$^\dagger$ uses a pretrained lane detector. The DET$_p$ scores for TopoNet, TopoMLP, and TopoLogic are computed using their official codebases. ``--'' denotes the absence of relevant data.}
\label{tab:main}
\small\setlength{\tabcolsep}{3.5pt}
\begin{tabular}{@{}llccccc>{\columncolor{gray!24}}cc@{}}
\toprule
Data & Method & Venue & DET$_l${\scriptsize$\uparrow$} & DET$_t${\scriptsize$\uparrow$} & TOP$_{ll}${\scriptsize$\uparrow$} & TOP$_{lt}${\scriptsize$\uparrow$} & OLS{\scriptsize$\uparrow$} & DET$_p${\scriptsize$\uparrow$} \\
\midrule
\multirow{6}{*}{\rotatebox{90}{\small sub-A}}
& TopoNet~\cite{li2023toponet} & arXiv'23 & 28.6 & 48.6 & 10.9 & 23.8 & 39.8 & 43.8 \\
& TopoMLP~\cite{wu2024topomlp} & ICLR'24 & 28.3 & 49.5 & 21.6 & 26.9 & 44.1 & 43.4 \\
& TopoLogic~\cite{fu2024topologic} & NeurIPS'24 & 29.9 & 47.2 & 23.9 & 25.4 & 44.1 & 45.2 \\
& TopoFormer$^\dagger$~\cite{lv2025topoformer} & CVPR'25 & \textbf{34.7} & 48.2 & 24.1 & 29.5 & 46.3 & -- \\
& TopoPoint~\cite{fu2025topopoint} & NeurIPS'25 & 31.4 & 55.3 & 28.7 & 30.0 & 48.8 & 52.6 \\
& \textbf{TopoCurve} & -- & 33.9 & \textbf{55.6} & \textbf{30.7} & \textbf{33.0} & \textbf{50.6} & \textbf{56.8} \\
\midrule
\multirow{6}{*}{\rotatebox{90}{\small sub-B}}
& TopoNet~\cite{li2023toponet} & arXiv'23 & 24.3 & 55.0 & 6.7 & 16.7 & 36.8 & 38.5 \\
& TopoMLP~\cite{wu2024topomlp} & ICLR'24 & 26.6 & 58.3 & 21.0 & 19.8 & 43.8 & 39.6 \\
& TopoLogic~\cite{fu2024topologic} & NeurIPS'24 & 25.9 & 54.7 & 21.6 & 17.9 & 42.3 & 39.2 \\
& TopoFormer$^\dagger$~\cite{lv2025topoformer} & CVPR'25 & \textbf{34.8} & 58.9 & 23.2 & 23.3 & 47.5 & -- \\
& TopoPoint~\cite{fu2025topopoint} & NeurIPS'25 & 31.2 & 60.2 & 28.3 & 27.1 & 49.2 & 45.1 \\
& \textbf{TopoCurve} & -- & 33.4 & \textbf{61.5} & \textbf{30.8} & \textbf{29.4} & \textbf{51.2} & \textbf{48.3} \\
\bottomrule
\end{tabular}
\end{table}

\subsection{Ablation Studies}
\label{sec:ablations}
\textbf{Effect of components.} All ablations use OpenLane-V2 subset\_A with TopoPoint as baseline.
Table~\ref{tab:incremental} shows the B\'{e}zier parameterization (B\'{e}z) builds the geometric foundation, improving lane detection (32.0 vs.\ 31.4 DET$_l$). FourierGeoBias (F) models directional alignment, boosting topology (30.1 vs.\ 28.9 TOP$_{ll}$). B\'{e}zier-Distributed Deformable Attention (BD) expands receptive fields across curves, enhancing endpoint accuracy (55.5 vs.\ 53.8 DET$_p$), while Curve-Anchored Redundancy Supervision (CA) densifies training to reach 50.6 OLS. Table~\ref{tab:leaveout} confirms removing the B\'{e}zier formulation causes the steepest drop (48.7 vs.\ 50.6 OLS), proving it anchors the geometric cascade.

\begin{table}[ht]
\centering
\begin{minipage}[t]{0.50\textwidth}
\centering
\caption{\textbf{Incremental ablation.}}
\label{tab:incremental}
\small\setlength{\tabcolsep}{1pt}
\begin{tabular}{@{}lcccccc>{\columncolor{gray!24}}cc@{}}
\toprule
B\'{e}z & F & BD & CA & DET$_l${\scriptsize$\uparrow$} & TOP$_{ll}${\scriptsize$\uparrow$} & TOP$_{lt}${\scriptsize$\uparrow$} & OLS{\scriptsize$\uparrow$} & DET$_p${\scriptsize$\uparrow$} \\
\midrule
 & & & & 31.4 & 28.7 & 30.0 & 48.8 & 52.6 \\
\checkmark & & & & 32.0 & 28.9 & 30.6 & 49.1 & 53.0 \\
\checkmark & \checkmark & & & 32.2 & 30.1 & 31.3 & 49.6 & 53.8 \\
\checkmark & \checkmark & \checkmark & & 33.2 & 30.2 & 32.0 & 50.1 & 55.5 \\
\checkmark & \checkmark & \checkmark & \checkmark & \textbf{33.9} & \textbf{30.7} & \textbf{33.0} & \textbf{50.6} & \textbf{56.8} \\
\bottomrule
\end{tabular}
\end{minipage}\hfill%
\begin{minipage}[t]{0.50\textwidth}
\centering
\caption{\textbf{Leave-one-out ablation.}}
\label{tab:leaveout}
\small\setlength{\tabcolsep}{1pt}
\begin{tabular}{@{}lccc>{\columncolor{gray!24}}ccc@{}}
\toprule
Config & DET$_l${\scriptsize$\uparrow$} & TOP$_{ll}${\scriptsize$\uparrow$} & TOP$_{lt}${\scriptsize$\uparrow$} & OLS{\scriptsize$\uparrow$} & DET$_p${\scriptsize$\uparrow$} & $\Delta$ \\
\midrule
Full & \textbf{33.9} & \textbf{30.7} & \textbf{33.0} & \textbf{50.6} & \textbf{56.8} & -- \\
w/o B\'{e}z & 31.4 & 28.5 & 29.8 & 48.7 & 53.0 & $-$1.9 \\
w/o F & 33.8 & 29.4 & 31.3 & 49.9 & 56.5 & $-$0.7 \\
w/o BD & 32.8 & 30.4 & 32.6 & 50.2 & 56.0 & $-$0.4 \\
w/o CA & 33.2 & 30.2 & 32.0 & 50.1 & 55.5 & $-$0.5 \\
\bottomrule
\end{tabular}
\end{minipage}
\end{table}

\textbf{Geometric encoding.} Table~\ref{tab:fourier} shows fusing raw distance (D) and tangent (T) brings additive improvements over distance or tangent alone (49.9--50.0 vs.\ 50.1 OLS). Table~\ref{tab:encoding} indicates positional, log-linear, and radial basis function encodings offer negligible gains over these scalars (50.2--50.3 vs.\ 50.1 OLS). Applying Fourier encoding to these fused inputs overcomes spectral bias, driving significant topology improvements (30.7 vs.\ 29.9 TOP$_{ll}$, 33.0 vs.\ 31.8 TOP$_{lt}$). Expanding Fourier bands from $L{=}4$ to $L{=}8$ yields finer discrimination, maximizing performance (50.6 vs.\ 50.4 OLS).\begin{table}[t]
\centering
\begin{minipage}[t]{0.51\textwidth}
\centering
\caption{\textbf{Geometric encoding comparison.}}
\label{tab:encoding}
\small\setlength{\tabcolsep}{0pt}
\begin{tabular}{@{}lcccc>{\columncolor{gray!24}}cc@{}}
\toprule
Encoding & DET$_l${\scriptsize$\uparrow$} & DET$_t${\scriptsize$\uparrow$} & TOP$_{ll}${\scriptsize$\uparrow$} & TOP$_{lt}${\scriptsize$\uparrow$} & OLS{\scriptsize$\uparrow$} & DET$_p${\scriptsize$\uparrow$} \\
\midrule
Raw & 33.8 & 55.5 & 29.9 & 31.8 & 50.1 & 56.5 \\
PE & 33.8 & 55.6 & 30.1 & 32.0 & 50.2 & 56.5 \\
RBF & 33.8 & 55.5 & 30.3 & 32.3 & 50.3 & 56.6 \\
Log-lin & 33.8 & 55.6 & 30.0 & 31.9 & 50.2 & 56.5 \\
Fourier & \textbf{33.9} & \textbf{55.6} & \textbf{30.7} & \textbf{33.0} & \textbf{50.6} & \textbf{56.8} \\
\bottomrule
\end{tabular}
\end{minipage}\hfill%
\begin{minipage}[t]{0.49\textwidth}
\centering
\caption{\textbf{FourierGeoBias analysis.}}
\label{tab:fourier}
\small\setlength{\tabcolsep}{0pt}
\begin{tabular}{@{}lcccc>{\columncolor{gray!24}}cc@{}}
\toprule
Config & DET$_l${\scriptsize$\uparrow$} & DET$_t${\scriptsize$\uparrow$} & TOP$_{ll}${\scriptsize$\uparrow$} & TOP$_{lt}${\scriptsize$\uparrow$} & OLS{\scriptsize$\uparrow$} & DET$_p${\scriptsize$\uparrow$} \\
\midrule
D, raw & 33.8 & 55.5 & 29.4 & 31.3 & 49.9 & 56.5 \\
T, raw & 33.8 & 55.6 & 29.6 & 31.5 & 50.0 & 56.5 \\
D+T, raw & 33.8 & 55.5 & 29.9 & 31.8 & 50.1 & 56.5 \\
D+T, F4 & 33.9 & 55.6 & 30.3 & 32.4 & 50.4 & 56.6 \\
D+T, F8 & \textbf{33.9} & \textbf{55.6} & \textbf{30.7} & \textbf{33.0} & \textbf{50.6} & \textbf{56.8} \\
\bottomrule
\end{tabular}
\end{minipage}
\end{table}

\textbf{Post-processing and efficiency.} Table~\ref{tab:postprocess} shows how TopoPoint relies on PLGM to reach 48.8 OLS. TopoCurve exceeds this natively. Applying PLGM to TopoCurve actually degrades performance (50.5 OLS), confirming our representation addresses endpoint deviation end-to-end.
Evaluated on a Tesla A100 GPU (Table~\ref{tab:efficiency}), TopoCurve yields a 3.69\% OLS gain over TopoPoint for only a 2.74\% GFLOPs increase, proving accuracy improvements outpace marginal costs. Despite richer geometry, it maintains compact parameters and real-time inference (13.1 FPS), ensuring deployability.

\subsection{Qualitative Results}
\label{sec:qualitative}
\begin{figure}[H]
\centering
\includegraphics[width=\textwidth]{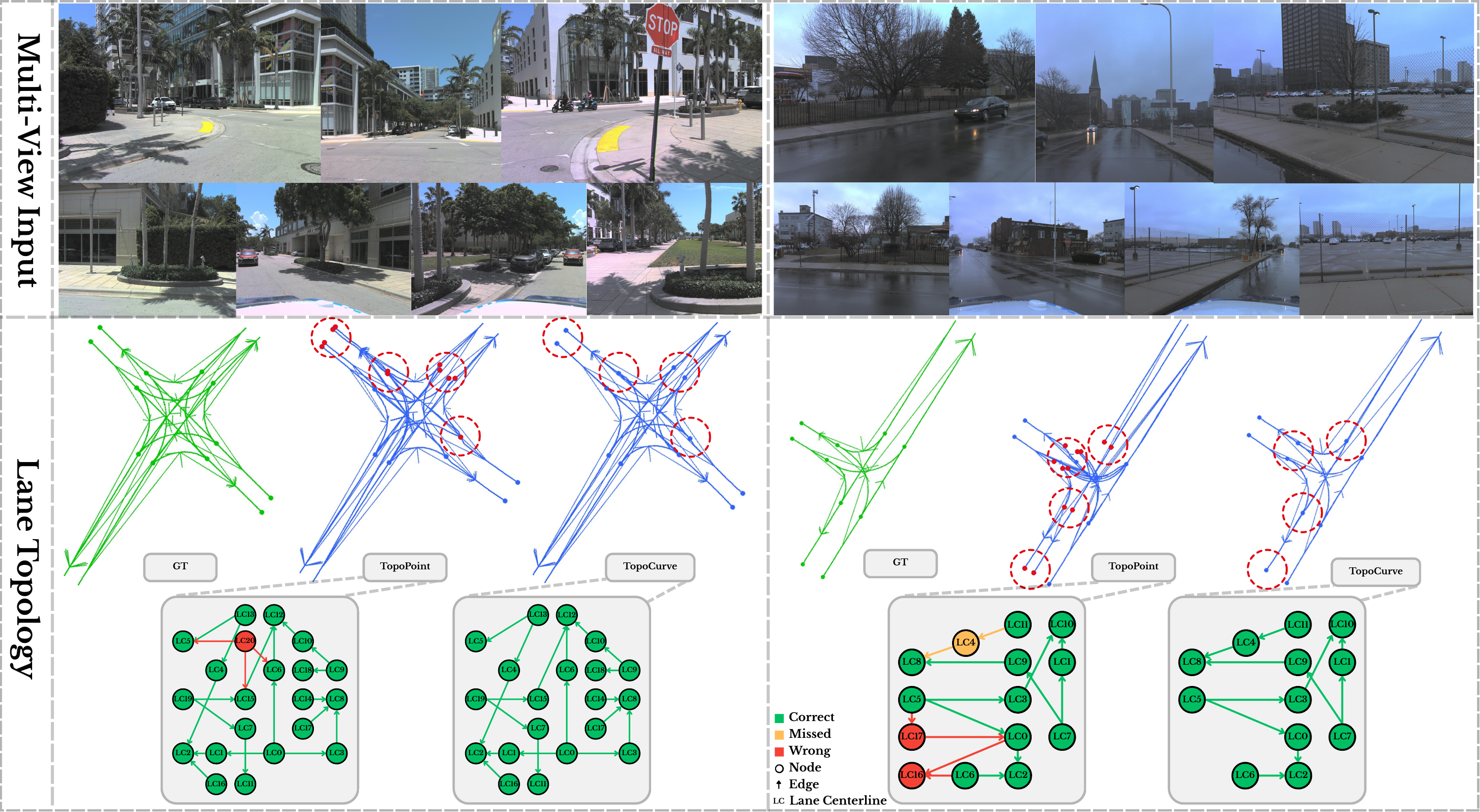}
\caption{\textbf{Qualitative comparison between TopoPoint and TopoCurve.} We showcase multi-view inputs, BEV lane detections, and the corresponding lane-lane topology graphs. TopoCurve consistently produces smoother curves, precise endpoint alignment, and more accurate topology.}
\label{fig:qualitative}
\end{figure}

Figure~\ref{fig:qualitative} provides qualitative comparisons between TopoCurve and TopoPoint. TopoPoint frequently produces multiple predictions for a single centerline, duplicate endpoints, jagged self-crossing lanes, and missed lanes or topologies. Furthermore, its PLGM heuristic often fails to snap distant endpoints or generates deformed lanes that appear artificially warped.
TopoCurve's continuous parametric representation yields smoother predictions, superior endpoint alignment and accurate topology, without relying on post-processing. Figures~\ref{fig:comparison1} and~\ref{fig:comparison2} provide additional qualitative comparisons.

\begin{table}[t]
\centering
\begin{minipage}[t]{0.54\textwidth}
\centering
\caption{\textbf{Post-processing effect.} PLGM~\cite{fu2025topopoint}.}
\label{tab:postprocess}
\small\setlength{\tabcolsep}{1.5pt}
\begin{tabular}{@{}llccc>{\columncolor{gray!24}}cc@{}}
\toprule
Method & PP & DET$_l${\scriptsize$\uparrow$} & TOP$_{ll}${\scriptsize$\uparrow$} & TOP$_{lt}${\scriptsize$\uparrow$} & OLS{\scriptsize$\uparrow$} & DET$_p${\scriptsize$\uparrow$} \\
\midrule
TopoPoint & -- & 30.8 & 28.0 & 29.2 & 48.3 & 51.8 \\
TopoPoint & PLGM & 31.4 & 28.7 & 30.0 & 48.8 & 52.6 \\
TopoCurve & -- & \textbf{33.9} & \textbf{30.7} & \textbf{33.0} & \textbf{50.6} & \textbf{56.8} \\
TopoCurve & PLGM & 33.7 & 30.6 & 33.0 & 50.5 & 56.8 \\
\bottomrule
\end{tabular}
\end{minipage}\hspace{0.02\textwidth}%
\begin{minipage}[t]{0.44\textwidth}
\centering
\caption{\textbf{Efficiency comparison (A100).}}
\label{tab:efficiency}
\small\setlength{\tabcolsep}{3pt}
\begin{tabular}{@{}lccc>{\columncolor{gray!24}}c@{}}
\toprule
Method & Params{\scriptsize$\downarrow$} & GFLOPs{\scriptsize$\downarrow$} & FPS{\scriptsize$\uparrow$} & OLS{\scriptsize$\uparrow$} \\
\midrule
TopoNet & 38.6M & 712.1G & 16.6 & 39.8 \\
TopoLogic & 37.8M & 665.0G & 17.1 & 44.1 \\
TopoPoint & 67.9M & 989.3G & 14.8 & 48.8 \\
TopoCurve & 79.5M & 1016.4G & 13.1 & \textbf{50.6} \\
\bottomrule
\end{tabular}
\end{minipage}
\end{table}

\section{Conclusion}
\label{sec:conclusion}
We present TopoCurve, a geometry-driven architecture that reparameterises lanes as endpoint-fixed cubic B\'{e}zier curves and systematically exploits this representation. Analytical tangents enable Fourier-encoded topology, curve samples anchor geometry-spanning deformable attention, and parallel attention blocks provide curve-anchored redundancy supervision.
This cascade achieves state-of-the-art results on OpenLane-V2 and surpasses TopoPoint on its proposed DET$_p$ metric, all without any post-processing, proving that endpoint deviation is addressed end-to-end.

\textbf{Limitations and future work.}
TopoCurve currently operates on single frames. Promising extensions include multi-frame reasoning, Frenet frame projections, and learned curve degree selection.

\textbf{Broader impact.}
TopoCurve advances topology reasoning for autonomous driving, but no perception system should be deployed in safety-critical settings without thorough validation.

\begin{ack}
This work was supported by the Important Project of Common European Interest on Microelectronics and Communication Technologies (IPCEI ME/CT) EuroDrives 2/RUE Nr.2/C9/I4.
\end{ack}

\bibliographystyle{plainnat}

\clearpage
\appendix

\section{Additional Qualitative Results}
\label{sec:appendix_qualitative}

\begin{figure}[htbp]
\centering
\includegraphics[width=\textwidth,height=0.76\textheight,keepaspectratio]{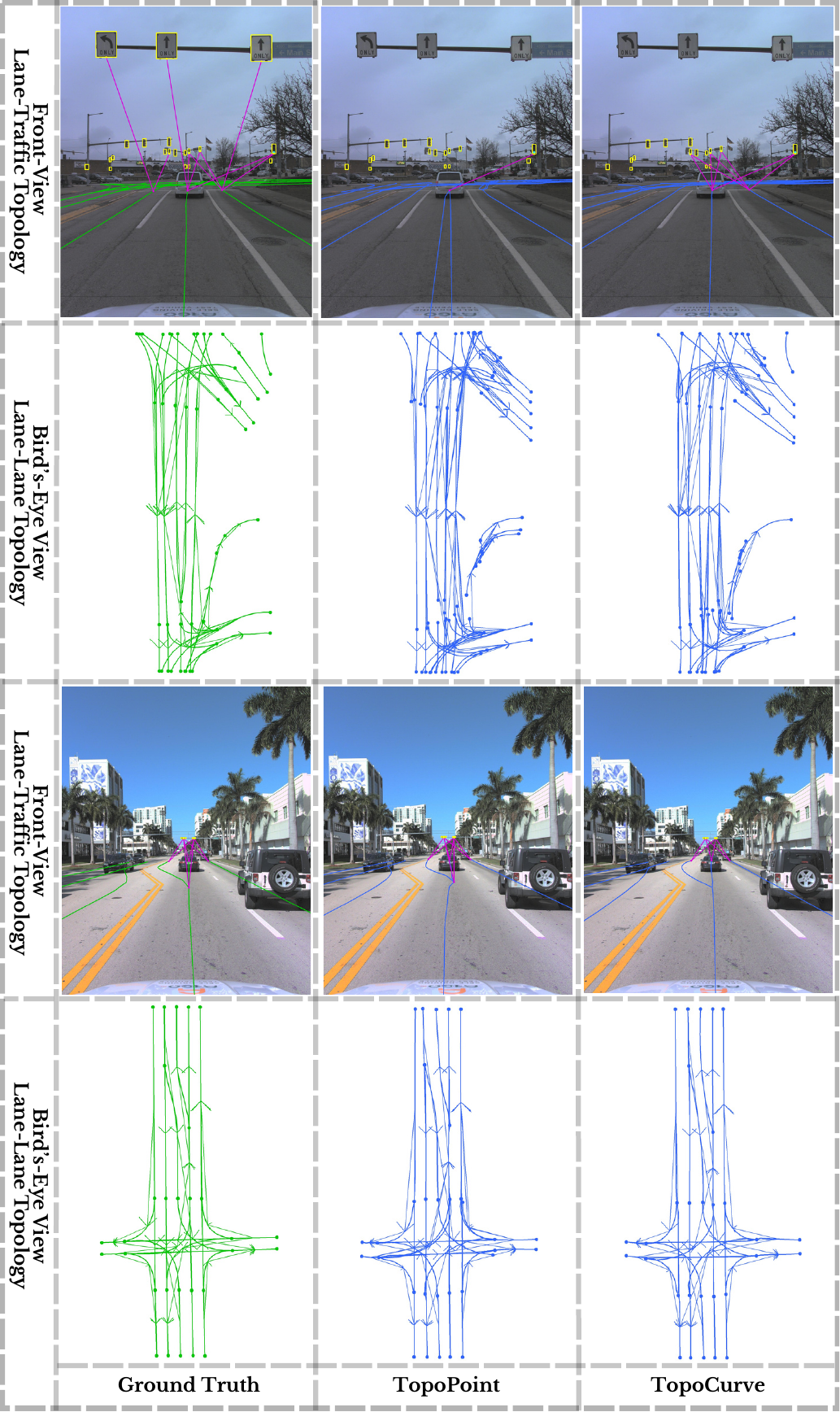}
\caption{\textbf{Additional qualitative comparison between TopoCurve and TopoPoint.} The visualization illustrates lane-to-traffic element topology in the front-view camera alongside lane-lane connectivity in the Bird's-eye view. TopoCurve consistently produces cleaner, smoother predictions, eliminating the jagged, self-intersecting artifacts prevalent in TopoPoint. Moreover, our continuous parametric representation achieves superior endpoint alignment and more accurate topological reasoning across both lane-lane and lane-traffic interactions.}
\label{fig:comparison1}
\end{figure}

\clearpage

\begin{figure}[htbp]
\centering
\includegraphics[width=\textwidth,height=0.76\textheight,keepaspectratio]{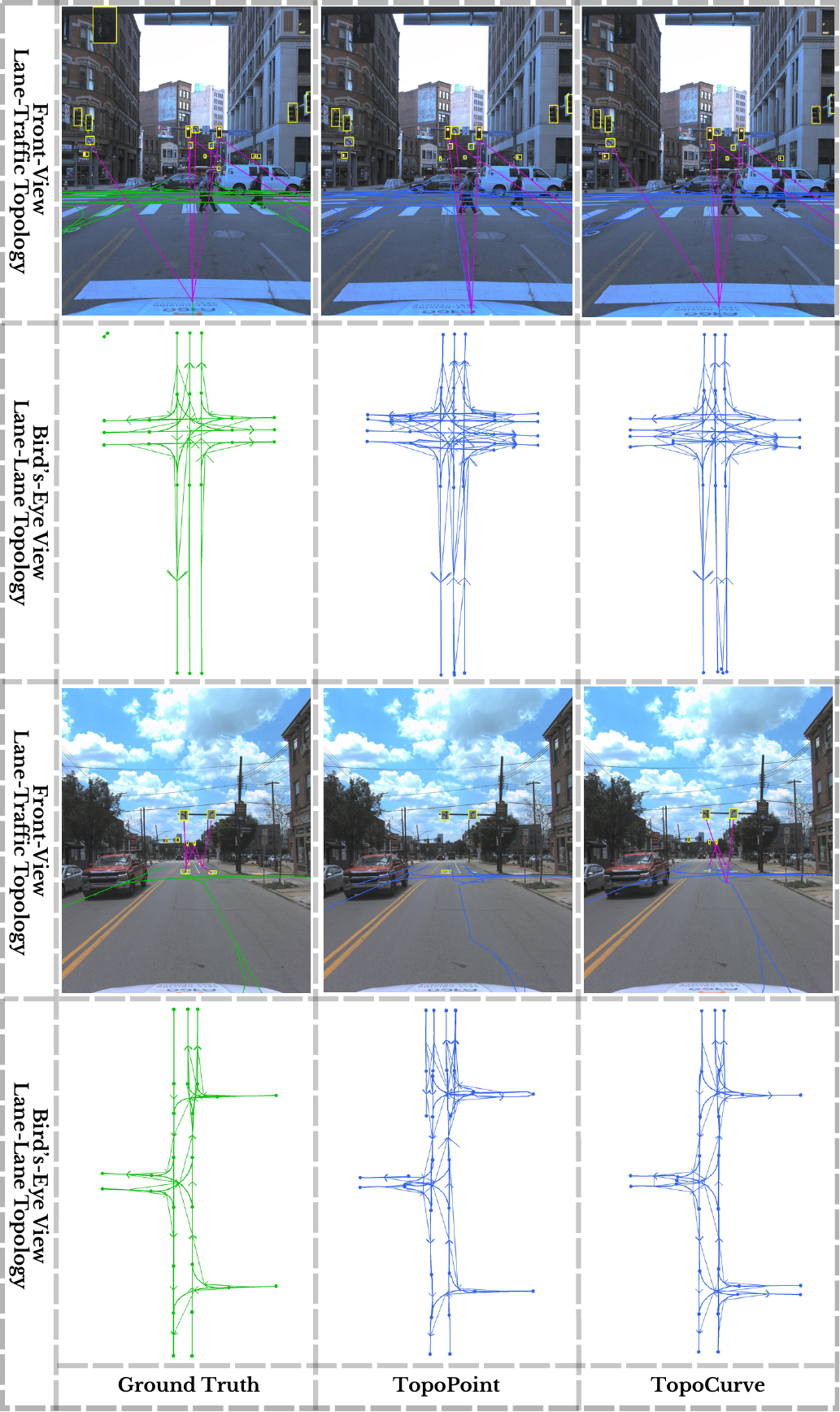}
\caption{\textbf{Additional qualitative comparison between TopoCurve and TopoPoint.} The visualization illustrates lane-to-traffic element topology in the front-view camera alongside lane-lane connectivity in the bird's-eye view. TopoCurve consistently produces cleaner, smoother predictions, eliminating the jagged, self-intersecting artifacts prevalent in TopoPoint. Moreover, our continuous parametric representation achieves superior endpoint alignment and more accurate topological reasoning across both lane-lane and lane-traffic interactions.}
\label{fig:comparison2}
\end{figure}

\clearpage
\section{Extended Architecture Details}
\label{sec:appendix_architecture}

This section elaborates on architectural components that are summarised in Section~\ref{sec:overview} of the main text, providing the level of detail necessary for full reproducibility.

\subsection{Decoder Layer Pipeline}
\label{sec:appendix_decoder}

Each of the $L{=}6$ decoder layers follows a fixed sequence: lane cross-attention, point cross-attention, joint self-attention, and graph-based feed-forward processing, interleaved with layer normalisation and residual connections.

\textbf{Lane and point cross-attention.}
Lane queries attend to BEV features through the $M{=}4$ parallel BDDA blocks described in Section~\ref{sec:bdda}. Their outputs are concatenated and linearly projected from $4d$ to $d$. Point queries independently attend to BEV features via standard multi-scale deformable attention~\cite{zhu2021deformable} with separate learned 3D reference points.

\textbf{Joint self-attention.}
To enable global information exchange, point and lane queries are concatenated into $\mathbf{Q}_{pl} = [\mathbf{Q}_p;\, \mathbf{Q}_l] \in \mathbb{R}^{(N_p + N_l) \times d}$ and processed by standard multi-head self-attention with 8 heads. The result is split back by index. TopoCurve omits the geometric attention bias that TopoPoint~\cite{fu2025topopoint} applies within self-attention, as geometric reasoning is concentrated within FourierGeoBias (Section~\ref{sec:fourier}).

\textbf{Graph-based feed-forward (USGN).}
The standard FFN is replaced by the Unified Scene Graph Network (Section~\ref{sec:appendix_usgn}), which enriches queries through graph convolution. After each layer, lane queries are decoded into intermediate B\'{e}zier control points for auxiliary supervision.

\subsection{Unified Scene Graph Network}
\label{sec:appendix_usgn}

The USGN~\cite{fu2025topopoint} augments the feed-forward stage with graph-based message passing. It receives point queries $\mathbf{Q}_p$, lane queries $\mathbf{Q}_l$, and traffic element queries $\mathbf{Q}_t$, along with adjacency matrices derived from the previous decoder layer (Section~\ref{sec:appendix_adjacency}).

Point and lane queries are first independently processed by two-layer FFNs (hidden dimension $2d{=}512$). The resulting features enter the first round of a bidirectional Point-Lane Graph Convolution (PLGCN), which simultaneously propagates lane information to points and point information to lanes via GCN~\cite{kipf2017gcn} message passing along the point-lane adjacency $\mathbf{A}_{pl}$. Each GCN maintains separate self-loop, forward, and backward weight matrices to model the asymmetric bipartite relationship, with a fixed edge weight $\alpha{=}0.6$.

After the first PLGCN round, lane features undergo two parallel GCN branches: a lane-lane GCN aggregates intra-lane context along $\mathbf{A}_{ll}$ using separate forward ($\mathbf{A}_{ll}$) and backward ($\mathbf{A}_{ll}^\top$) message passing, while a lane-traffic GCN aggregates traffic element features along $\mathbf{A}_{lt}$, using $C_t{=}13$ class-conditioned weight matrices weighted by detached per-class sigmoid scores. The outputs of both branches are concatenated and linearly projected from $2d$ to $d$.

A second PLGCN round with independent parameters operates on the updated representations, and the final output is combined with the USGN input via residual connections.

\subsection{Geometric Adjacency Computation}
\label{sec:appendix_adjacency}

The adjacency matrices used by the USGN are recomputed at each decoder layer from the current (detached) predictions and passed to the next layer. All adjacencies are initialised to zero before layer 0.

\textbf{Lane-lane adjacency} ($\mathbf{A}_{ll}$) captures endpoint-to-startpoint proximity. For lanes $i$ and $j$, we compute the L1 distance $D^{ll}_{ij} = \|\hat{\mathbf{L}}_i[K{-}1] - \hat{\mathbf{L}}_j[0]\|_1$ and convert it to a soft adjacency via a learnable kernel:
\begin{equation}
    A^{ll}_{ij} = \exp\!\Bigl(-\frac{(D^{ll}_{ij})^P}{w}\Bigr) \cdot (1 - \delta_{ij})
\end{equation}
where $P$ and $w$ are learnable scalars (initialised to 2 and 10), and self-loops are excluded.

\textbf{Point-lane adjacency} ($\mathbf{A}_{pl}$) blends geometric proximity with predicted topology. The minimum L1 distance from point $k$ to either endpoint of lane $j$ is $D^{pl}_{kj} = \min(\|\hat{P}_k - \hat{\mathbf{L}}_j[0]\|_1,\; \|\hat{P}_k - \hat{\mathbf{L}}_j[K{-}1]\|_1)$. The adjacency combines a distance kernel with the topology head prediction:
\begin{equation}
    A^{pl}_{kj} = \lambda_1 \cdot \sigma\!\bigl(\hat{G}^{pl}_{kj}\bigr) + \lambda_2 \cdot \exp\!\Bigl(-\frac{(D^{pl}_{kj})^{P'}}{w'}\Bigr)
\end{equation}
where $\lambda_1, \lambda_2$ are learnable mixing weights (initialised to 1) and $P', w'$ are separate learnable kernel parameters (initialised to 2 and 10).

\textbf{Lane-traffic adjacency} ($\mathbf{A}_{lt}$) is taken as the sigmoid of the lane-traffic topology prediction from the previous decoder layer.

\subsection{Traffic Element Detector}
\label{sec:appendix_te}

Traffic elements are detected by a standard Deformable DETR~\cite{zhu2021deformable} head operating on front-view multi-scale features. The detector comprises a 6-layer encoder and 6-layer decoder with iterative bounding box refinement, $N_t{=}100$ object queries, and $C_t{=}13$ traffic element classes. Its detection loss combines focal loss~\cite{lin2017focal} for classification, L1 for regression, and GIoU loss~\cite{rezatofighi2019giou}. Before entering the USGN at each decoder layer, traffic element features are projected through a per-layer embedding network (two-layer MLP with hidden dimension $2d$), producing representations compatible with the graph convolution modules.

\subsection{Query Initialization}
\label{sec:appendix_queries}

TopoCurve initialises $N_l{=}300$ lane queries, $N_p{=}200$ point queries, and $N_t{=}100$ traffic element queries as learned embeddings of dimension $d{=}256$. Lane and point reference points are learned 3D positions in normalised BEV coordinates spanning $[-51.2, 51.2] \times [-25.6, 25.6] \times [-2.3, 1.7]$\,m. At layer 0, the $K{=}11$ curve references for BDDA are initialised by replicating the 2D projection of each lane reference point, reducing BDDA to standard single-point deformable attention. From layer 1 onward, curve references are computed from the predicted B\'{e}zier control points via the Bernstein basis: $\mathbf{p}^{(i)}_{1:K} = \mathbf{A}\hat{\mathbf{C}}^{(i-1)}$. Point reference points are updated independently at each layer via their regression head.

\end{document}